\documentclass[11pt]{article}
\usepackage[utf8]{inputenc}       % 第二段有，第一段缺（编码统一）
\usepackage{authblk}              % 必须放在xcolor/graphicx之前（关键！）
\usepackage{setspace}             % 控制行距（\onehalfspacing依赖此包）
\usepackage{xcolor}
\usepackage{graphicx}
\usepackage{booktabs}
\usepackage{multirow}
\usepackage[margin=1in]{geometry} % 第二段是1.25in，若要完全一致可改为1.25in
\usepackage{amsmath}
\usepackage{amssymb}
\usepackage{parskip}              % Nature风格段落间距（第二段隐含配置）
\usepackage[numbers,sort&compress]{natbib}
\usepackage{hyperref}             % 必须放在最后（关键！）

\title{Logos: An evolvable reasoning engine for rational molecular design}
\author[1]{Haibin Wen}     
\author[1]{Zhe Zhao}    
\author[4]{Fanfu Wang}    
\author[1]{Tianyi Xu}        
\author[3]{Hao Zhang}       
\author[2]{Chao Yang}      
\author[1]{Ye Wei}   
\affil[1]{City University of Hong Kong, Hong Kong 999077, China}         
\affil[2]{Shanghai Jiao Tong University, Shanghai 200240, China} 
\affil[3]{Riltide Medicines, Beijing, China}                      
\affil[4]{Lanzhou University, Lanzhou 730000, China}    

\date{}
\begin{document}

\maketitle

\begin{abstract}
The discovery and design of functional molecules remain central challenges across chemistry, biology, and materials science. While recent advances in machine learning have accelerated molecular property prediction and candidate generation, existing models tend to excel either in physical fidelity without transparent reasoning, or in flexible reasoning without guarantees of chemical validity. This imbalance limits the reliability of artificial intelligence systems in real scientific design workflows.

Here we present Logos, a compact molecular reasoning model that integrates multi-step logical reasoning with strict chemical consistency. Logos is trained using a staged strategy that first exposes the model to explicit reasoning examples linking molecular descriptions to structural decisions, and then progressively aligns these reasoning patterns with molecular representations. In a final training phase, chemical rules and invariants are incorporated directly into the optimization objective, guiding the model toward chemically valid outputs.

Across multiple benchmark datasets, Logos achieves strong performance in both structural accuracy and chemical validity, matching or surpassing substantially larger general-purpose language models while operating with a fraction of their parameters. Beyond benchmark evaluation, the model exhibits stable behaviour in molecular optimization tasks involving multiple, potentially conflicting constraints. By explicitly exposing intermediate reasoning steps, Logos enables human inspection and assessment of the design logic underlying each generated structure. These results indicate that jointly optimizing for reasoning structure and physical consistency offers a practical pathway toward reliable and interpretable AI systems for molecular science, supporting closer integration of artificial intelligence into scientific discovery processes.
\end{abstract}

\section{Introduction}

% The discovery of functional molecules underpins progress across a wide range of scientific disciplines, including drug development, materials science, and energy research\cite{reymond2015chemical,vamathevan2019applications,tabor2018accelerating}. Central to this challenge is the vastness of chemical space: even when restricted to structures that obey basic chemical stability and synthetic feasibility constraints, the number of potentially accessible molecules is estimated to exceed $10^{60}$\cite{ruddigkeit2012gdb}. Within such an immense space, traditional discovery paradigms, which are largely driven by expert intuition and incremental trial-and-error, struggle to scale, increasingly limiting the pace of innovation\cite{butler2018machine,sanchez2018inverse}.

Molecular discovery is a central driver of advances in pharmaceutical development, materials engineering, and sustainable energy\cite{reymond2015chemical,vamathevan2019applications,tabor2018accelerating}. The theoretical space of synthetically accessible, stable molecules is immense-estimated at $10^{60}$\cite{ruddigkeit2012gdb} or more, yet exploration of this chemical space is severely constrained by traditional paradigms that rely on human intuition and iterative empiricism \cite{butler2018machine,sanchez2018inverse}. Overcoming the resulting scalability limits and high attrition rates requires a paradigm shift. In particular, the field needs advanced artificial intelligence architectures that not only generate molecular candidates but also rationalize them, creating a feedback loop in which computational proposals are systematically evaluated and refined by scientific expertise\cite{joshi2021artificial,zenil2026future,lee2025rethinking}.

% Machine learning has emerged as a powerful tool to address this challenge, enabling high-throughput prediction of molecular properties and accelerating the generation of candidate structures\cite{olivecrona2017reinvent,survey2024genai}. Deep learning architectures, particularly Transformers and graph neural networks (GNNs)\cite{gilmer2017mpnn,jensen2019gnn,you2022graph}, have established strong performance in mapping between molecular representations and physicochemical or biological properties\cite{stokes2020deep,schuster2019transfer}. These advances have substantially reduced the cost of screening and optimization for well-defined objectives. However, many real-world scientific tasks extend beyond statistical prediction or pattern matching. They require the ability to interpret abstract, often hierarchical design constraints and to translate them into concrete molecular modifications while respecting strict physical and chemical rules\cite{bilodeau2022generative,fang2023mol,m2024augmenting}.
To facilitate this collaborative feedback loop, the AI must bridge human intuition and chemical reality. Meanwhile, molecular discovery is often posed as a generative problem: translating abstract linguistic specifications, such as human-desired properties, functions, or structural motifs, into concrete candidate structures. Because design objectives are largely articulated in text, inverse design can be viewed as the mapping from these semantic inputs to chemical entities\cite{edwards2022molt5,su2023moleculestm}. 
Although machine learning methods, especially transformers and graph neural networks\cite{gilmer2017mpnn,jensen2019gnn,you2022graph}, have achieved strong performance in molecular representation and property prediction\cite{olivecrona2017reinvent,survey2024genai,stokes2020deep,schuster2019transfer}, a critical gap remains. Many realistic design tasks require more than statistical pattern matching; they demand nuanced interpretation of complex, hierarchical constraints and their translation into valid molecular structures that obey physicochemical principles. This motivates a key question: can we engineer a model that systematically decomposes abstract targets into discrete structural modifications, ranging from scaffold selection to functional group optimization, rather than relying on generic textual inference alone\cite{bilodeau2022generative,fang2023mol,m2024augmenting}?

The language understanding and multi-step reasoning capabilities of large language models (LLMs) are well aligned with these requirements\cite{wei2022cot,ouyang2022instructgpt}. LLMs can interpret textual design briefs and approximate chemists’ reasoning from abstract properties to concrete structures, while few-shot and in-context learning address the sparsity and heterogeneity of molecular datasets\cite{liu2024moleculargpt}. This suggests the possibility of a direct ``description $\rightarrow$ structure'' workflow with auditable reasoning, enabling effective human--AI collaboration. However, current LLMs generally lack explicit chemical grounding and often produce syntactically plausible but chemically invalid structures, limiting their reliability in design pipelines\cite{irwin2022chemcrow,du2024molllm,white2023assessment}. Conversely, specialized scientific models achieve high chemical validity and domain accuracy but are opaque and poorly suited to natural language inputs. The central challenge, therefore, is to endow LLMs with robust chemical validity while preserving and exposing their reasoning, thereby combining language-native interaction with accountable, scientifically grounded molecular design\cite{zhang2024chemllm}.

% Together, these limitations expose a critical gap in current scientific AI systems: existing models tend to excel either at physical fidelity without reasoning transparency, or at flexible reasoning without guarantees of chemical validity. Bridging this gap is essential for enabling AI systems that can participate meaningfully in scientific design processes rather than merely generating candidate structures.

Here we introduce Logos, a compact molecular reasoning model designed to integrate multi-step logical reasoning with strict chemical consistency. Moving beyond mere scaling, Logos achieves expert-level reasoning within a controlled parameter budget by aligning training with scientific logic. It explicitly translates complex instructions into chemically valid and logically traceable molecular designs. To achieve this, we adopt an iterative, staged training strategy that progressively aligns linguistic reasoning with molecular structure\cite{edwards2021text2mol,su2023moleculestm}. First, to overcome the scarcity of explicit reasoning data, we augment existing description--structure pairs with intermediate reasoning steps generated by a larger language model. Next, supervised fine-tuning\cite{ouyang2022instructgpt} aligns these reasoning patterns with molecular representations for principled structure generation. Finally, instead of relying on external post hoc filters, chemical validity is internalized through a physics-guided reinforcement learning process\cite{shao2024grpo,obrien2023curriculum,horwood2023transformer}. By incorporating chemical rules evaluated by standard cheminformatics toolkits\cite{rdkit,heller2015inchi}, this final stage encourages the model to favor reasoning trajectories that naturally yield chemically valid molecules while suppressing invalid pathways. Crucially, the optimized model can then be deployed to further expand the reasoning dataset, driving subsequent cycles of this training pipeline.

% To achieve this, we adopt a staged training strategy that progressively aligns linguistic reasoning with molecular structure\cite{edwards2021text2mol,su2023moleculestm}. In the first stage, we address the scarcity of explicit reasoning data in molecular science by augmenting existing description--structure pairs with intermediate reasoning steps generated by a larger language model. These augmented examples provide the model with concrete illustrations of how abstract descriptions can be decomposed into chemically meaningful decisions. In the second stage, supervised fine-tuning\cite{ouyang2022instructgpt} aligns these reasoning patterns with molecular representations, enabling the model to follow structured instructions and to construct candidate molecules in a principled manner.

% Crucially, in the final stage, chemical validity is enforced directly through a physics-guided optimization process\cite{shao2024grpo,obrien2023curriculum,horwood2023transformer}. Instead of relying solely on textual likelihoods, the model is trained using reinforcement learning objectives that incorporate chemical rules and invariants, such as those evaluated by established cheminformatics toolkits\cite{rdkit,heller2015inchi}. This stage encourages the model to favour reasoning trajectories that lead to chemically consistent structures, while systematically suppressing pathways that produce invalid molecules. As a result, chemical constraints are internalised as part of the model's generation behaviour rather than applied as an external post hoc filter.

Across multiple benchmark datasets\cite{edwards2022molt5,preuer2018fcd}, Logos achieves strong performance in both structural accuracy and chemical validity, matching or exceeding much larger general-purpose models while operating at a fraction of their scale. Beyond benchmark metrics, the model exhibits stable behaviour in multi-constraint molecular optimization tasks, where competing objectives must be balanced through iterative structural refinement. Logos exposes its intermediate reasoning steps, allowing researchers to inspect and evaluate the logic underlying each design decision.

\begin{figure}[htbp]
    \centering
    \includegraphics[width=\textwidth, height=0.675\textheight, keepaspectratio]{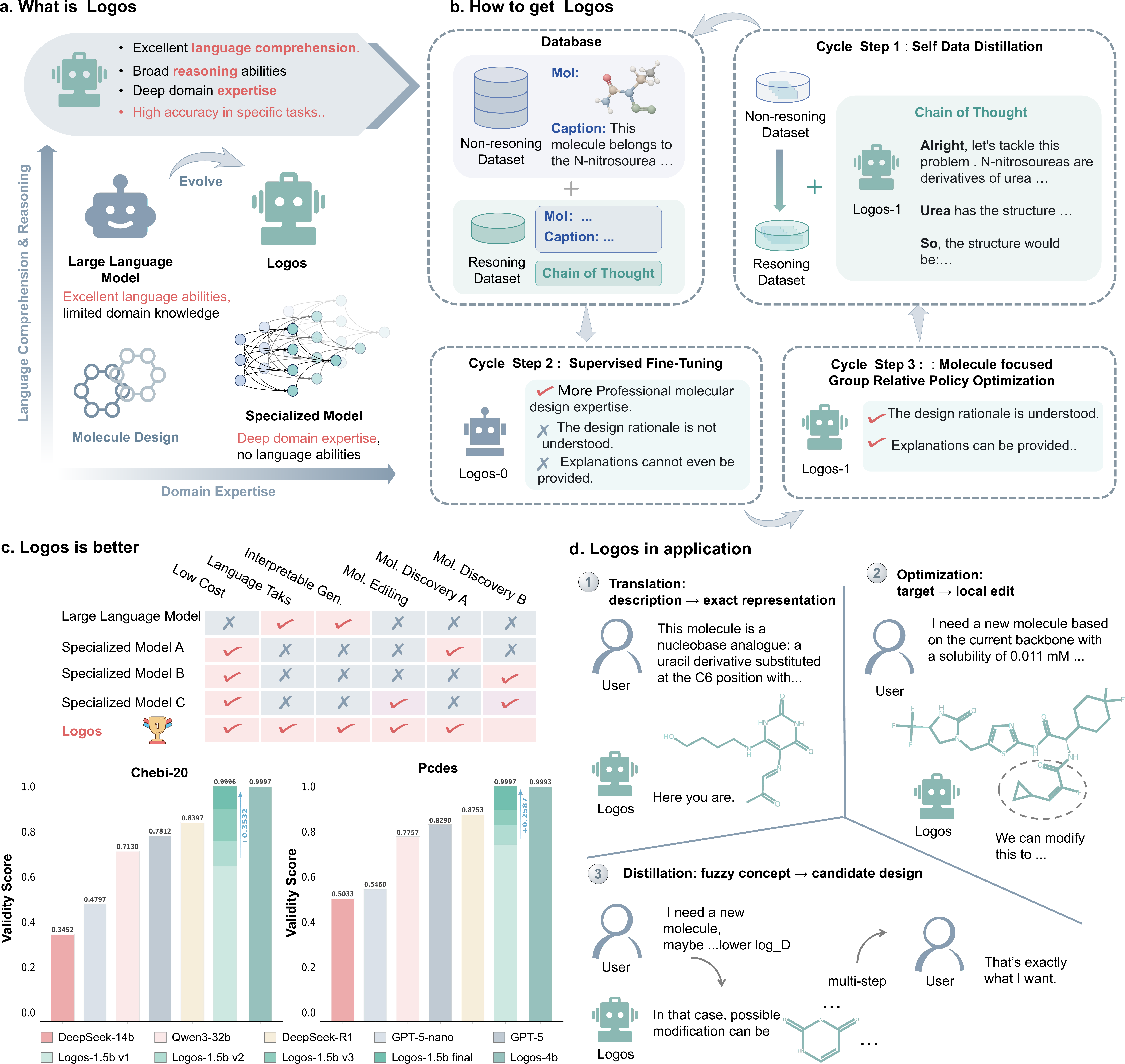}
    \caption{Conceptual framework and training pipeline of Logos. 
    \textbf{a,} Logos combines the chemical accuracy of specialized models (bottom lright) with the reasoning of general LLMs (top left), addressing both interpretability and structural validity. 
    \textbf{b,} Three-stage pipeline: Cycle 1 (self-data distillation) builds chain-of-thought (CoT) data with a teacher model; Cycle 2 (supervised fine-tuning) trains on molecular design; Cycle 3 (molecule-focused GRPO) uses reinforcement learning with chemical rewards. 
    \textbf{c,} Validity scores on ChEBI-20 and PCdes; Logos-1.5b (final) and Logos-4b approach $\sim$99.9\%, outperforming larger baselines. 
    \textbf{d,} Logos enables interactive molecular design through three paradigms}
    \label{fig:framework}
\end{figure}

Taken together, these results suggest that reliable scientific AI systems need not sacrifice interpretability for performance, nor physical consistency for reasoning flexibility. By jointly optimizing for logical structure and chemical validity, Logos illustrates a pathway toward AI models that function as transparent, trustworthy collaborators in molecular science. In the following sections we present the system architecture and training pipeline (Fig.~\ref{fig:framework}), benchmark results across caption-to-molecule tasks and the evolutionary training mechanism (Fig.~\ref{fig:benchmarks}, Fig.~\ref{fig:evolutionary_framework}), and the application of Logos to interactive, multi-objective molecular design (Fig.~\ref{fig:integrated_workflows}). We then discuss trade-offs, limitations, scale versus reasoning, and future work (Discussion). Full details of the task format, data preparation, supervised and reinforcement training, and bootstrapping procedure are given in Methods.
\section{Results}

\subsection{Motivation and Three-Stage Training Framework}

Current molecular design approaches divide into two categories with complementary weaknesses (Fig.~\ref{fig:framework}a). Specialized models such as GNN-based generators achieve high chemical accuracy but cannot accept free-text design briefs or explain their decisions in natural language. General-purpose LLMs articulate multi-step reasoning but frequently output chemically invalid SMILES because they are not grounded in valency or topological constraints. The former handle tasks where inputs and outputs are both structural or numerical, but fail when presented with open-ended instructions. The latter follow complex prompts fluently yet produce structures that violate basic chemical rules, limiting their reliability in workflows where validity is required. Logos is designed to combine the chemical fidelity of specialized models with the reasoning transparency of LLMs, so that both the final structure and the logic underlying it can be examined and revised.

The model is trained in three stages (Fig.~\ref{fig:framework}b). In Cycle 1 (self-data distillation), a larger teacher model generates chain-of-thought (CoT) reasoning for existing molecule--caption pairs, producing a dataset that pairs descriptions with explicit reasoning steps and structural outputs. Standard chemical databases provide only caption--structure pairs; this step synthesises the missing reasoning layer. The student then learns to reproduce both the reasoning style and the structural output, preparing it for the reinforcement stage in which validity is directly rewarded. In Cycle 2 (supervised fine-tuning, SFT), a 1.5B-parameter student is trained on these data and becomes an intermediate model, Logos-0, capable of following instructions and outputting both reasoning and a molecule. In Cycle 3 (molecule-focused group relative policy optimization, GRPO), reinforcement learning rewards encode chemical validity via RDKit checks and consistency with the ground truth. SFT alone does not guarantee that valency or other physical constraints are satisfied at generation time; the reward signal provides a direct incentive for validity. These three cycles operate iteratively, progressively augmenting the original molecule--caption dataset with high-quality CoT reasoning paths. A more capable 4B-parameter model is subsequently trained on this final dataset, undergoing both SFT and GRPO in a single unified pass.

On ChEBI-20 and PCdes\cite{edwards2021text2mol,edwards2022molt5}, Logos-1.5b (final) reaches validity scores of 0.9996 and 0.9997, and Logos-4b achieves similarly near-perfect performance (Fig.~\ref{fig:framework}c), above all larger baselines in the figure. Fig.~\ref{fig:framework}d illustrates the intended use pattern: the user specifies scaffold and property constraints; the model returns a candidate molecule with its reasoning in a \texttt{<think>} block; 
Logos supports diverse discovery workflows: exact description-to-molecule translation, targeted scaffold optimization, and multi-step conceptual distillation. In the latter, the user can supply validation feedback (e.g., from experiments or simulations), and the model uses this to adjust its reasoning and strategy in subsequent rounds, making the system suitable for iterative design rather than single-shot generation. The performance gains afforded by this framework are quantified in the sections that follow.
\subsection{Superior Performance across Multi-Dataset Benchmarks}

\begin{figure}[htbp]
    \centering
     \includegraphics[width=\textwidth, height=0.75\textheight, keepaspectratio]{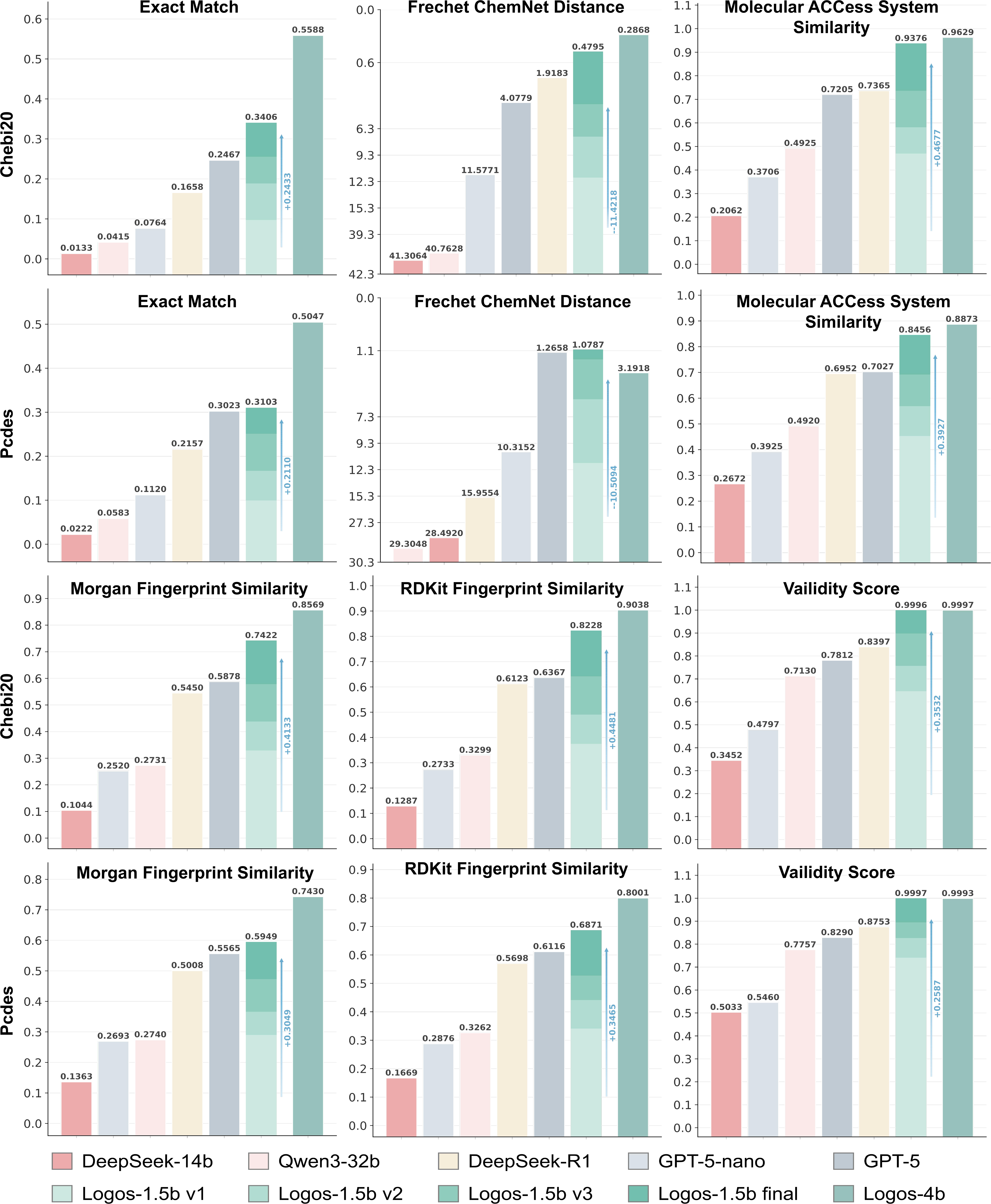}    
    \caption{Benchmarking across Logos-1.5b versions (v1 to final), Logos-4b and general LLMs (e.g., DeepSeek-14b, GPT-5) on ChEBI-20 and PCdes. Logos-1.5b final reaches exact match 0.3406 (ChEBI-20) and validity $\sim$1.0. Structural similarity (MACCS, RDKit, Morgan) improves across versions; FCD decreases to 0.4795, indicating that generated molecules lie closer to real drug-like chemistry.}
    \label{fig:benchmarks}
\end{figure}

We evaluated Logos and comparison models on two caption-to-molecule benchmarks: ChEBI-20\cite{edwards2022molt5,hastings2016chebi}, which uses biochemical and functional descriptions, and PCdes, which uses physicochemical property descriptions. 
The two datasets probe complementary aspects of generalization, namely biological role versus property constraints, and together assess whether performance holds across different types of molecular design tasks.
We compared the full evolutionary series of Logos-1.5b (from initial version v1 through the final post-GRPO model) and the final Logos-4b against four general-purpose LLMs: DeepSeek-14b\cite{deepseek2025r1}, Qwen3-32b\cite{yang2025qwen3}, DeepSeek-R1\cite{deepseek2025r1}, and GPT-5 (Fig.~\ref{fig:benchmarks}). 

The validity score measures the fraction of generated SMILES satisfying basic chemical rules such as valency\cite{preuer2018fcd}. 
General-purpose models consistently score below 1.0: GPT-5 reaches 0.7812 on ChEBI-20 and 0.8290 on PCdes, while DeepSeek-R1 achieves 0.8397 and 0.8753, respectively. 
Logos-1.5b (final) reaches 0.9996 and 0.9997, with invalid structures largely eliminated after the GRPO stage. 
Exact match (EM) measures graph-level identity with the ground-truth molecule. 
Logos-1.5b (final) achieves EM 0.3406 on ChEBI-20 and 0.3103 on PCdes; Logos-4b improves these to 0.5588 and 0.5047, compared with 0.2467 and 0.3023 for GPT-5. 
The gain from v1 to the final 1.5b model is +0.2433 on ChEBI-20 and +0.2110 on PCdes, with the largest improvements concentrated at the GRPO stage, consistent with the reward signal directly shaping generation toward correct and valid structures. 
On PCdes the same ordering holds: Logos (final) leads both on validity and EM over the general-purpose baselines. 

To capture partial correctness and structural proximity, we computed fingerprint-based similarities (MACCS, RDKit path, and Morgan\cite{rdkit,preuer2018fcd}). % [cite: 1]
MACCS reflects the presence of key functional groups; on ChEBI-20, Logos-1.5b (final) reaches 0.9376 and Logos-4b improves this to 0.9629, versus 0.7365 for DeepSeek-R1 and 0.7205 for GPT-5. % [cite: 1]
RDKit and Morgan similarities reach 0.8228 and 0.7422 for Logos-1.5b (final), and an impressive 0.9038 and 0.8569 for Logos-4b, versus 0.6367 and 0.5878 for GPT-5 on ChEBI-20, indicating that the generated molecules align well with the targets in topology and local atomic environment. % [cite: 1]
The Fréchet ChemNet distance (FCD)\cite{preuer2018fcd} measures the distance between the distribution of generated molecules and a reference set of drug-like molecules in a learned feature space; a lower FCD corresponds to more realistic, drug-like distributions; we use the same metric and reference set as in prior work for comparability. % [cite: 1]
Logos-1.5b (final) reaches FCD 0.4795 on ChEBI-20, while Logos-4b achieves an even lower FCD of 0.2868, versus 4.0779 for GPT-5 and 1.9183 for DeepSeek-R1, and FCD drops sharply from v1 to the final 1.5b model on both datasets (Fig.~\ref{fig:benchmarks}). % [cite: 1]
Taken together, the 4B-parameter Logos outperforms the much larger general-purpose models we tested on validity, exact match, structural similarity, and distributional realism, indicating that molecule-focused training with chemical rewards can compensate for smaller scale on these tasks. % [cite: 1]
% ==========================================
% Figure 2 Place Holder
% ==========================================
\subsection{Evolutionary Training Mechanism, Ablation Analysis, and Structured Reasoning Format}

\begin{figure}[htbp]
    \centering
    \includegraphics[width=\textwidth,height=0.85\textheight,keepaspectratio]{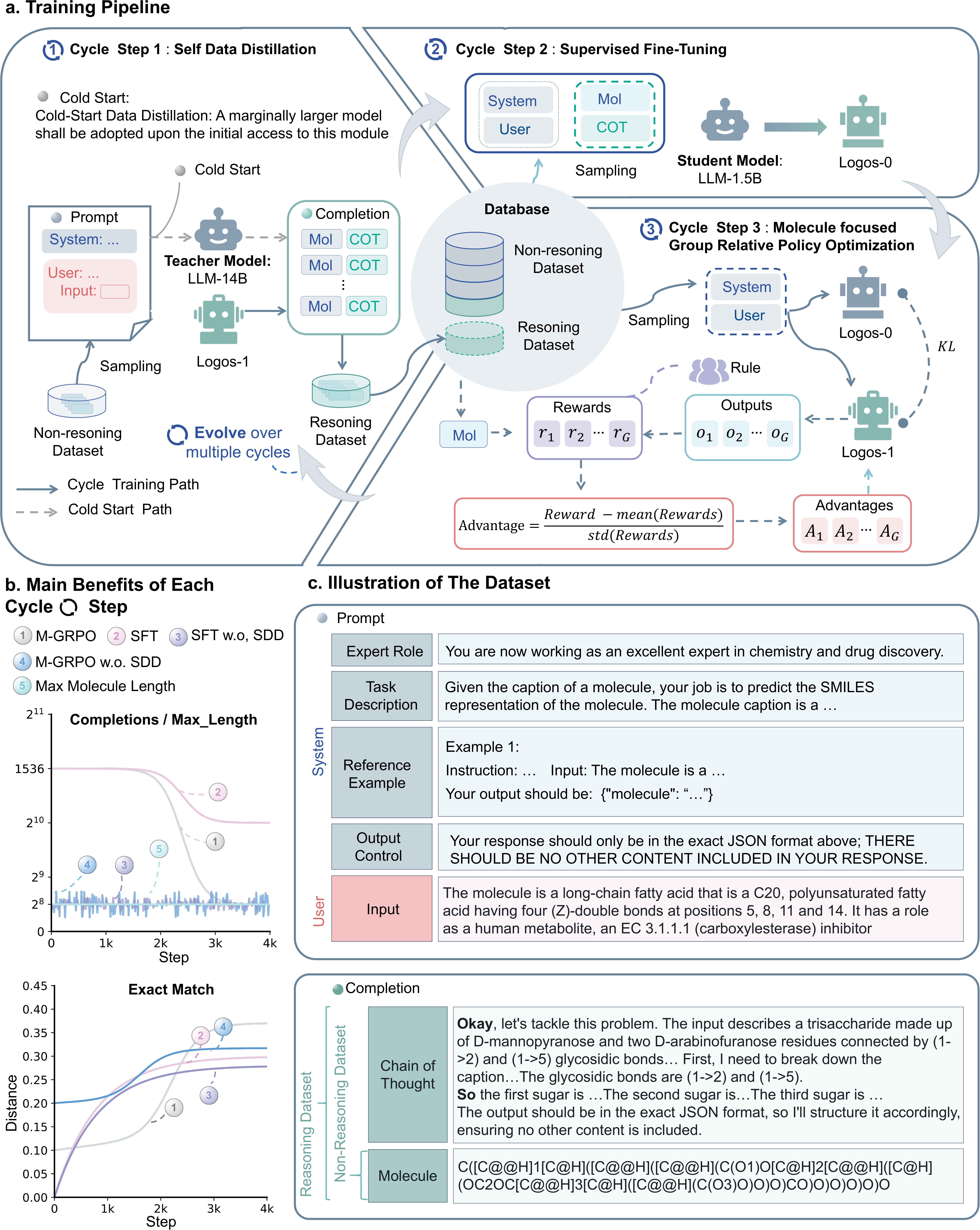} % Placeholder
    \caption{Evolutionary training and output format. 
    \textbf{a,} Pipeline: teacher (LLM-14B) generates CoT for description--structure pairs (Cycle 1); student (LLM-1.5B) is fine-tuned on CoT data to become Logos-0 (Cycle 2); M-GRPO with chemical rewards yields Logos (Cycle 3). 
    \textbf{b,} Ablations: full M-GRPO (solid blue) reaches EM $>$ 0.35; removing self-data distillation (w.o. SDD) lowers performance; M-GRPO stabilizes generation length. 
    \textbf{c,} Output format: system prompt plus JSON; the model outputs reasoning in \texttt{<think>} then the molecule in JSON, e.g. for a trisaccharide task.}
    \label{fig:evolutionary_framework}
\end{figure}

The training procedure is organized into three cycles that are summarized in Fig.~\ref{fig:evolutionary_framework}a. Molecular databases typically provide caption--structure pairs without explicit reasoning steps\cite{wei2022cot,edwards2022molt5,hastings2016chebi}. To obtain a training signal for reasoning, we first use a larger teacher model (14B parameters) to generate chain-of-thought (CoT) for these pairs; the teacher is prompted to explain how the description maps to structural decisions before outputting the SMILES. The resulting CoT dataset is then used to train a smaller student (1.5B/4B parameters). In Cycle 2, the student is supervised on this dataset and becomes the intermediate model Logos-0, which can follow instructions and output both a reasoning block and a molecule. Supervised training alone, however, does not guarantee that generated molecules are chemically valid\cite{olivecrona2017reinvent,ronan2024reinvent4}. In Cycle 3 we apply molecule-focused group relative policy optimization (M-GRPO)\cite{shao2024grpo}: for each prompt the model produces multiple completions; each completion receives a scalar reward that encodes chemical validity (e.g., valency checks), match or similarity to the ground truth, and structural fingerprint similarity; a KL penalty limits drift from the initial policy. The policy is updated using the relative advantage of each completion within its group, so that trajectories that yield valid, correct molecules are favoured. This stage yields the final Logos model.

The benchmark results above reflect the full three-stage pipeline; ablation experiments reported here isolate the contribution of each stage and examine how the output format itself underpins both the training signal and downstream interpretability.
When the full M-GRPO pipeline is used, including self-data distillation in Cycle 1, exact match on the evaluation set exceeds 0.35 (Fig.~\ref{fig:evolutionary_framework}b). Removing self-data distillation, whether under M-GRPO or SFT alone, reduces performance to approximately 0.20--0.25, indicating that the CoT data produced in Cycle 1 cannot be replaced by caption--structure pairs alone. Maximum generation length over training further distinguishes the regimes: under M-GRPO, generation length converges to a stable range, whereas the SFT-only model produces longer and often redundant outputs, suggesting that the reward signal suppresses unfocused reasoning chains. These ablations support the use of all three cycles in the final pipeline.

All training and evaluation use a fixed output format (Fig.~\ref{fig:evolutionary_framework}c). The model must first emit a reasoning block bounded by \texttt{<think>} and \texttt{</think>} tags, then a single JSON object whose \texttt{molecule} field contains the SMILES. The system prompt defines the task and, when applicable, includes few-shot caption--molecule examples. {For a complex request, such as generating a trisaccharide with specified glycosidic linkages, the model states the linkage logic and stereochemical constraints in the reasoning block before producing the SMILES, explicitly tying each structure to a verifiable reasoning trace.} This format serves two purposes: it makes chemical validity checkable by tools such as RDKit, enabling the GRPO reward computation, and it makes the design logic readable by a human reviewer. Requiring the reasoning block to precede the answer also reduces the risk of the model short-circuiting to a bare SMILES output, which would undermine both auditability and iterative refinement. This explicit reasoning structure directly supports the human-in-the-loop application described in the next section.
% ==========================================
% Figure 4 Caption
% ==========================================
\begin{figure}[htbp]
    \centering
    \includegraphics[width=\textwidth]{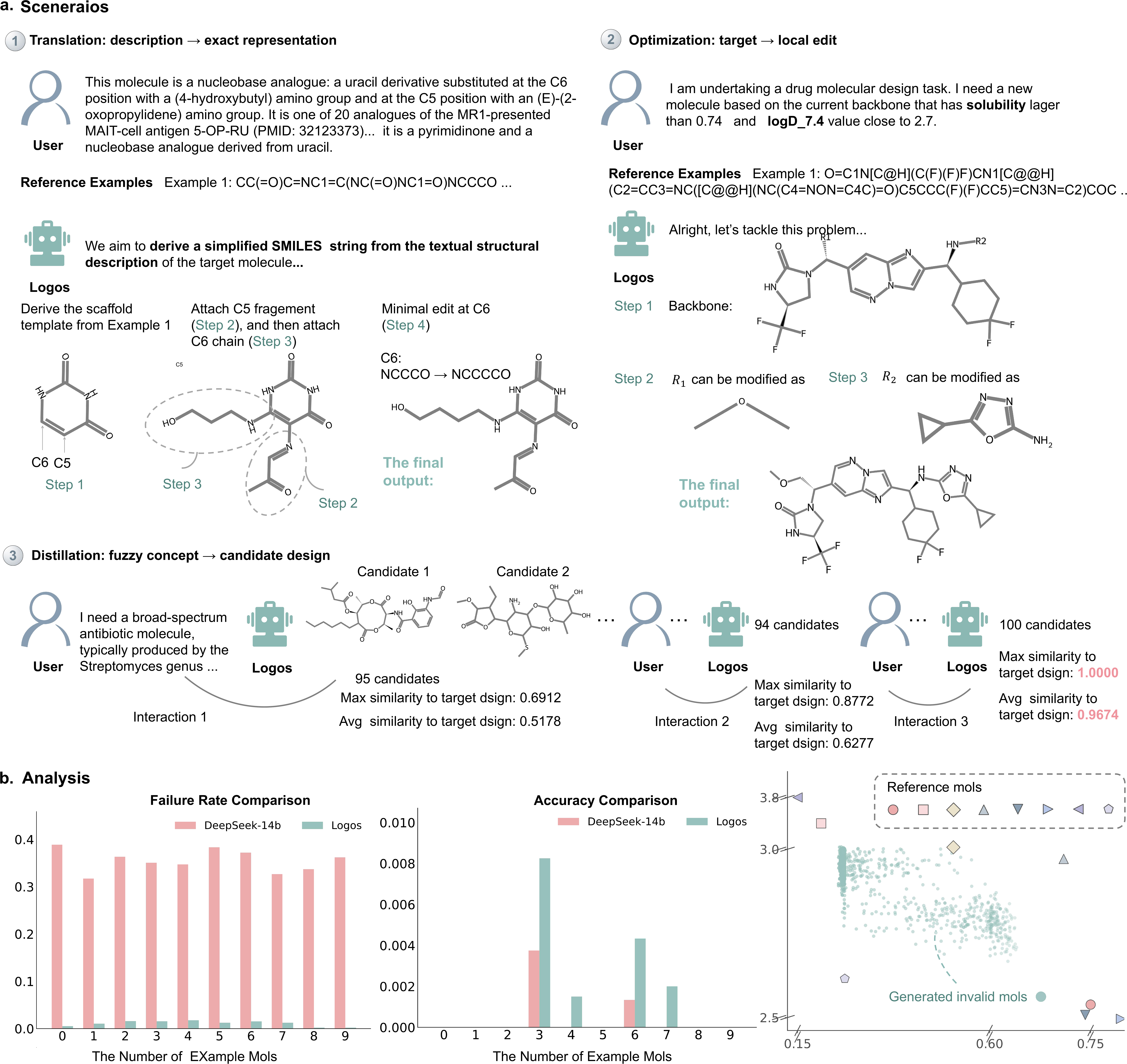} % Placeholder
    \caption{Interactive Application to Multi-Objective Molecular Optimization. 
    \textbf{a,} Three exemplary paradigms for rational molecule design: (1) \textbf{Translation}, where the model converts a detailed textual description into an exact molecular representation through step-by-step structural assembly; (2) \textbf{Optimization}, which performs local edits on a given backbone to satisfy specific physicochemical property constraints (e.g., $\log D_{7.4}$ and solubility); and (3) \textbf{Distillation}, an iterative multi-turn process that refines candidate molecules from a fuzzy conceptual query, gradually converging on the desired molecule.
    % \textbf{b,} CoT breakdown: scaffold identification, site localization, and group modifications show that outputs follow a structured reasoning path. 
    % \textbf{c,} Task success in multi-parameter optimization: DeepSeek-r1 (14B) has higher validation rate (98.75\%); Logos (4B) reaches 64.46\% with fewer parameters. 
    % \textbf{d,} Generated molecules cluster near the target $\log D_{7.4}$ under solubility and scaffold constraints.
    \textbf{b.} Analysis of Logos' performance in real-world cases of multi-objectice  optimization from~\cite{velcicky2024discovery}.
    }
    \label{fig:integrated_workflows}
\end{figure}

\subsection{Interactive Application to Multi-Objective Molecular Optimization}

Benchmark evaluation addresses performance on well-defined tasks, such as caption-to-molecule generation and molecule editing. Figure~\ref{fig:integrated_workflows}a (1 and 2) illustrates how Logos processes these foundational tasks. Building on this, Figure~\ref{fig:integrated_workflows}a (3) demonstrates how users can initiate a query with vague requirements, draw inspiration from the model's explicit chain of thought, and interactively refine their criteria to obtain the desired molecule. Furthermore, we now examine whether this explicit reasoning capability transfers to iterative, multi-objective optimization closer to real-world practice. In lead optimization, chemists cycle between design hypotheses and experimental or in silico validation, balancing objectives, such as $\log D_{7.4}$, solubility, and scaffold retention, that frequently conflict\cite{waring2010lipophilicity,d2012multi}. We tested Logos under a human-in-the-loop protocol and compared it with a larger general-purpose reasoning model (Fig.~\ref{fig:integrated_workflows}b). $\log D_{7.4}$ and solubility were chosen as exemplar constraints because they are routinely paired in drug discovery and can work against each other: raising lipophilicity tends to reduce aqueous solubility, so the model must navigate a trade-off rather than optimize a single quantity.

 The user specifies constraints (for example, retain a given scaffold, achieve desired $\log D_{7.4}$ and solubility), and the model returns candidate molecules with its reasoning in the \texttt{<think>} block. The user then provides feedback from assays or property predictions, and the model incorporates this in the next round. Because the reasoning is visible, the user can identify which structural choices drove each candidate and redirect the logic; black-box generators that output only a structure leave the user to infer what to change, making iteration slower and less targeted.

% Fig.~\ref{fig:integrated_workflows}b illustrates how the chain-of-thought decomposes into inspectable steps: scaffold identification from the request or in-context examples, localization of modification sites, and functional group substitution by analogy. On multi-parameter tasks requiring both validity and constraint satisfaction, DeepSeek-r1 (14B parameters) attains a validation rate of 98.75\% on the evaluated prompts, while Logos (4B parameters) reaches 64.46\% with roughly one-third of the parameters (Fig.~\ref{fig:integrated_workflows}c). Under the stated constraints, Logos's candidates cluster near the target $\log D_{7.4}$ value (Fig.~\ref{fig:integrated_workflows}d), demonstrating that the model steers generation toward the specified region of property space even when several constraints are simultaneously active. Logos thus supports iterative, multi-objective molecular design workflows in which the reasoning behind each candidate is explicit and open to inspectiond.
The first two panels of Fig.~\ref{fig:integrated_workflows}b illustrate the molecule generation failure rate (i.e., invalid molecules) and the success rate of generating molecules with the correct $ \log D_{7.4}$  and solubility values when DeepSeek-14B and Logos are provided with varying numbers of reference samples containing explicit property values. In comparison, Logos generates a higher proportion of chemically valid molecules and achieves a greater success rate in generating molecules with the desired $\log D_{7.4}$ and solubility values. Furthermore, when the user expects a $ \log D_{7.4}$ of 2.8 and a solubility of 0.56, the third panel of Fig.~\ref{fig:integrated_workflows}b presents a similarity comparison between the generated valid molecules, whose actual $\log D_{7.4}$ and solubility cannot be pre-determined, and known samples. While the actual properties of these molecules remain unverified, structural similarity analysis indicates that they closely resemble known samples with the target profiles. This demonstrates that the model steers generation toward the specified region of property space even when several constraints are simultaneously active. Logos thus supports iterative, multi-objective molecular design workflows in which the reasoning behind each candidate is explicit and open to inspectiond.
\section{Discussion}

The results above support two main conclusions: that a compact, domain-trained model can match or exceed much larger general-purpose LLMs on caption-to-molecule benchmarks when reasoning is combined with explicit chemical constraints, and that the same model can participate in iterative, human-in-the-loop design when multiple objectives must be balanced. Here we discuss the trade-offs implied by these findings, the limitations of the current work, the relationship between scale and reasoning in molecular tasks, and directions for future work.

 \paragraph{Trade-offs.}
The advantages of Logos in accuracy and interpretability introduce two primary trade-offs. First, there is a trade-off between inference latency and auditability. Traditional black-box generators, such as graph neural networks (GNNs) or diffusion models, as well as standard LLMs operating without a chain-of-thought, can generate molecular structures rapidly. In contrast, Logos must autoregressively decode an explicit, often lengthy \texttt{<think>} reasoning trace before outputting a SMILES string. This generates significantly more tokens per request, trading raw single-pass generation speed for high physical consistency and human-verifiable logic. Consequently, Logos is exceptionally well-suited for precision-driven, human-in-the-loop workflows, but less optimal for massive high-throughput virtual screening where microsecond latency is critical. Second, achieving state-of-the-art molecular reasoning with a compact model (e.g., 4B parameters) necessitates a trade-off between domain specialization and general applicability. To match or exceed the chemical accuracy of massive general-purpose models, Logos allocates its limited parameter budget almost entirely to chemical syntax, spatial reasoning, and physicochemical alignment. Therefore, while it excels as a specialized domain expert, it inevitably sacrifices the broad, cross-disciplinary knowledge  inherent to larger foundation models.

\paragraph{Limitations.}
Several limitations should be borne in mind. First, we evaluated on two caption-to-molecule benchmarks and one multi-parameter optimization protocol; performance on other tasks (e.g., retrosynthesis, reaction prediction, or other property targets) may differ. Second, the quality of the reasoning produced by the teacher in Cycle 1 and by Logos at inference time is not directly measured; we rely on validity and structural match as proxies. Third, the model is trained and evaluated in English; captions in other languages or mixed-language settings were filtered out and are not represented. Fourth, the bootstrapping step depends on the current model being able to generate correct molecules for some previously failed prompts; if the failure rate is very high, the gain from bootstrapping may be small. Fifth, we did not compare against all existing molecular generators (e.g., diffusion or VAE-based models\cite{vignac2022digress,gomez2018automatic}) on the same benchmarks; our comparison is focused on general-purpose LLMs and on the Logos evolutionary series. Finally, the human-in-the-loop scenario was illustrated with one constraint set ($\log D_{7.4}$, solubility, scaffold); behaviour under other constraints or with different numbers of feedback rounds has not been systematically varied.

\paragraph{Scale versus reasoning.}
A recurring question in scientific AI is whether scaling model size is necessary for strong performance or whether domain-specific training can compensate for smaller scale. On the caption-to-molecule benchmarks we report, the 4B Logos (final) outperforms the much larger general-purpose models on validity, exact match, and distributional realism. This suggests that for this task, molecule-focused training with chemical rewards can offset the advantage of scale. The result is consistent with the view that general-purpose models are not optimised for chemical validity and that injecting physical constraints via the reward signal narrows the gap. It does not imply that scale is irrelevant: the teacher model used for CoT distillation is larger (14B). Extending this comparison to other model sizes and to other scientific domains would clarify the generality of this trade-off.

\paragraph{Future work.}
Several directions follow naturally. (i) \emph{Scaling the student:} training a larger student (e.g., 7B or 14B) with the same pipeline could show whether the gains from molecule-focused training persist at larger scale and whether multi-parameter task success improves. (ii) \emph{Benchmarks and tasks:} evaluating on additional caption-to-molecule datasets, on retrosynthesis or reaction prediction, and on property-based design with explicit objectives would strengthen the evidence for generalisation. (iii) \emph{Reasoning quality:} developing metrics or human evaluations for the correctness and usefulness of the generated reasoning (rather than only the molecule) would allow the community to optimise and compare interpretability. (iv) \emph{Integration with experiments:} coupling Logos with synthesis planning or experimental feedback loops (e.g., closing the loop with assay or property data) would test its value in real discovery campaigns. (v) \emph{Bootstrapping and data:} scaling the bootstrapping phase, incorporating active learning, or curating higher-quality CoT from experts could improve data efficiency and final performance. (vi) \emph{Other modalities:} extending the format to include spectra, reactions, or other chemical representations would broaden the scope of the approach. We hope that the pipeline and the results reported here provide a basis for these and related efforts.

\section{Methods}

We implement a multi-stage pipeline that first distils chain-of-thought (CoT) reasoning from a larger teacher into a compact student, then refines the student using reinforcement learning whose rewards explicitly encode chemical validity and consistency with ground-truth structures (Fig.~\ref{fig:framework}). An optional bootstrapping step reuses the current model to generate high-quality reasoning for previously failed samples, expanding the training set without manual annotation. The three main stages, including data preparation and CoT distillation (Cycle 1), supervised reasoning training (Cycle 2), and molecule-focused policy optimization (Cycle 3), are summarised in Fig.~\ref{fig:framework}b and Fig.~\ref{fig:evolutionary_framework}. A central design principle is that a single, strict output format is used everywhere: data construction, supervised training, reward computation, and bootstrapping all assume the same structure, so that parsing and chemical validation can be applied uniformly and multiple reward terms can be combined in one reinforcement phase. Below we describe the task and output format, data construction, supervised training, the reinforcement phase with its reward design, and the bootstrapping loop; implementation details are given at the end.

\noindent\textbf{Task and output format.}

We consider the caption-to-molecule (C2M) task\cite{edwards2022molt5,edwards2021text2mol}: given a natural-language description of a molecule (e.g., physicochemical properties, functional role, or structural hints), the model must output the corresponding SMILES string\cite{weininger1988smiles}. This setting is representative of inverse design and knowledge-guided generation in drug and chemical discovery\cite{segler2018planning,survey2024genai}. A central design choice is to require the model to emit a reasoning trace before the answer: the model first produces text inside a dedicated block delimited by \texttt{<think>} and \texttt{</think>}, then a single JSON object containing the key \texttt{molecule} with the SMILES. 
% Parsing is performed by splitting the completion on the end of the reasoning block (e.g., \texttt{</think>} followed by newline), taking the second segment, and extracting the JSON; the \texttt{molecule} field is then validated and compared using RDKit\cite{rdkit} (e.g., validity, InChI\cite{heller2015inchi}, structural fingerprints). 
Formally, let $y$ denote the raw model completion. The parsing process is defined as a sequential extraction pipeline:
\begin{equation}
\hat{s} = \Phi(E_{\text{regex}}(a)) \quad \text{where} \quad y = [r; d; a]
\label{eq:parsing_pipeline}
\end{equation}
Here, $y$ is first split at the reasoning delimiter $d$ (e.g., \texttt{</think>\textbackslash n\textbackslash n}) to separate the reasoning trace $r$ from the raw answer segment $a$. The function $E_{\text{regex}}$ isolates the JSON payload via regular expression matching, and $\Phi$ serves as a robust string parser that yields the predicted SMILES $\hat{s}$. To ensure fault tolerance against formatting hallucinations, $\Phi$ incorporates format correction heuristics (e.g., brace and quote removal) and fallback key routing (checking \texttt{molecule}, \texttt{smiles}, etc.). Finally, $\hat{s}$ is mapped to a computational molecular graph $\hat{m}$:
\begin{equation}
    \hat{m} = \mathcal{V}(\hat{s})
\end{equation}
where $\mathcal{V}$ represents the strict chemical validation function powered by RDKit \cite{rdkit}. If $\hat{s}$ violates chemical valency rules, $\mathcal{V}$ rejects it ($\hat{m} = \emptyset$). The successfully parsed graph $\hat{m}$ is subsequently used for downstream programmatic evaluations, including InChI-based \cite{heller2015inchi} exact matching and structural fingerprint comparisons.
This separation allows the same pipeline to (i) train and evaluate interpretable, step-by-step reasoning, and (ii) apply programmatic chemical checks and multi-term rewards to the parsed molecule. The format is used consistently in data construction, supervised fine-tuning, reward computation, and distillation, so that no separate parsing logic is needed for each stage. Using a single, strict format across the pipeline also makes it possible to combine many reward terms in the reinforcement phase: each term receives the same parsed molecule and can call the same validation and fingerprint routines, which simplifies implementation and keeps reward definitions aligned with the data contract.

\noindent\textbf{Data preparation.}

% We start from existing molecule--caption pairs drawn from chemical databases\cite{hastings2016chebi,edwards2022molt5}. Each record includes the caption, the ground-truth molecule, and optional in-context (few-shot) caption--molecule examples used to build the system prompt. For each sample, the prompt is assembled from a task description and a variable number of few-shot examples; the model is instructed to predict the SMILES for the caption in the user message, supporting both zero-shot and few-shot training and evaluation. Some records are marked according to whether a high-quality CoT and correct molecule were previously obtained (e.g., from an earlier training run or distillation pass); this marking is used later for stratified sampling in the reinforcement set.
We start from an existing dataset of molecule--caption pairs $\mathcal{D} = \{(c_i, m_i^*)\}_{i=1}^N$ drawn from chemical databases \cite{hastings2016chebi,edwards2022molt5}, where $c_i$ represents the natural-language caption and $m_i^*$ is the ground-truth SMILES string. Each record optionally includes a set of $K$ in-context (few-shot) examples $E_i = \{(c_k, m_k)\}_{k=1}^K$ (where $K \ge 0$) used to build the system prompt. For each sample, the full input prompt $x_i$ is assembled by concatenating a fixed task description $\mathcal{T}$, the few-shot examples $E_i$, and the target caption $c_i$ in the user message, which can be formalized as $x_i = [\mathcal{T}; E_i; c_i]$. This formulation inherently supports both zero-shot ($K=0$) and few-shot ($K>0$) training and evaluation. Furthermore, each record is assigned a success indicator $q_i \in \{0, 1\}$ denoting whether a high-quality reasoning trace (CoT) and a correct molecule were previously generated for $c_i$ (e.g., from an earlier training run or distillation pass). The tuple $(K, q_i)$ serves as the condition variable for stratified sampling when constructing the reinforcement learning set later.

% For supervised reasoning training, we retain only samples that already have a high-quality CoT and a correct molecule, and we filter out reasoning traces that contain non-English characters. Each training example is a triple: system (task plus few-shot examples), user (target caption), and assistant (reasoning block followed by JSON with the ground-truth molecule). This yields the CoT dataset used to train the student.
For supervised reasoning training, we construct a highly curated dataset $\mathcal{D}_{\text{SFT}} \subseteq \mathcal{D}$. Specifically, we retain only samples where the success indicator $q_i = 1$ (i.e., possessing a high-quality CoT $r_i$ and the correct molecule $m_i^*$), and we apply a linguistic filter to discard reasoning traces containing non-English characters. Each training example is then structured as a conversational triple $\tau_i = (p_{\text{sys}}, p_{\text{user}}, p_{\text{asst}})$, defined as follows:
\begin{equation}
     p_{\text{sys}} = [\mathcal{T}; E_i], \quad p_{\text{user}} = c_i, \quad p_{\text{asst}} = [r_i; d; f_{\text{JSON}}(m_i^*)] 
\end{equation}
where $d$ is the structural delimiter (e.g., \texttt{</think>}) and $f_{\text{JSON}}(\cdot)$ denotes the function that wraps the ground-truth SMILES into the required JSON schema. This yields the final CoT dataset $\mathcal{D}_{\text{SFT}} = \{\tau_i\}_{i=1}^{N_{\text{SFT}}}$ used to train the student model via standard autoregressive language modeling.

% For the reinforcement phase, we construct a separate set of prompts that contain only the system and user parts (no assistant answer). Each of these is associated with the ground-truth molecule and, when applicable, the list of in-context example molecules. These fields are passed to the reward functions at training time so that each completion can be scored for validity, match to the ground truth, and whether the generated molecule is distinct from the in-context examples. Building the RL set from the same source as the SFT set (but with assistant answers omitted) keeps the two stages aligned: the model sees the same types of prompts and few-shot configurations in both supervised and reinforcement phases. To balance difficulty and few-shot configuration, we use stratified sampling: we partition samples by the number of few-shot examples and by whether the sample had previously yielded a correct CoT or not, then sample from each stratum. This ensures that the policy is trained on both easier and harder cases and across different in-context settings. We find that without this stratification, the reinforcement phase can become biased toward already-solved or single-shot prompts; with it, the model is exposed to a mix that supports few-shot-to-zero-shot generalization and more stable learning. The number of few-shot examples per sample can vary (e.g., zero, one, or several); stratifying by this number as well as by success/failure ensures that the policy is trained across the range of in-context settings that may appear at test time.
For the reinforcement phase, we construct a separate training set $\mathcal{D}_{\text{RL}}$ containing only the prompt contexts $x_i = (p_{\text{sys}}, p_{\text{user}})$, deliberately omitting the assistant's completion $p_{\text{asst}}$. Each prompt $x_i$ is internally coupled with a reference context $\mathcal{C}_i = (m_i^*, M_{E_i})$, where $M_{E_i} = \{m_k\}_{k=1}^K$ denotes the set of target molecules from the in-context examples. During RL training, $\mathcal{C}_i$ is hidden from the model but passed to the reward environment to evaluate the generated molecule $\hat{m}$. Specifically, the reward function scores the completion based on chemical validity, exact structural match against $m_i^*$, and generative novelty (i.e., ensuring $\hat{m} \notin M_{E_i}$). Building the RL set from the same source as the SFT set (but with assistant answers omitted) keeps the two stages aligned: the model sees the same types of prompts and few-shot configurations in both supervised and reinforcement phases.
To maintain a balanced distribution of difficulty and context length, we employ stratified sampling based on the previously defined criterion $(K, q_i)$. We partition the dataset into disjoint strata $\mathcal{S}_{K, q} = \{ x_i \in \mathcal{D} \mid |E_i| = K \land q_i = q \}$ and sample proportionally from each stratum. This explicit stratification ensures that the policy is trained across a diverse spectrum of easier ($q_i=1$) and harder ($q_i=0$) cases, as well as varying in-context settings ($K \ge 0$). We observe that without this partitioning, the RL phase tends to become biased toward already-solved or single-shot prompts. By enforcing this stratification, the model is exposed to a robust mixture that stabilizes the learning dynamics and strongly promotes few-shot-to-zero-shot generalization at test time.

\noindent\textbf{Supervised reasoning training.}

% The student is a 4B-parameter language model that natively supports the reasoning block used for CoT. We perform full-parameter fine-tuning on the CoT dataset, maximising the likelihood of the full assistant reply (reasoning plus JSON). To encourage substantive rather than trivial reasoning, we down-weight loss on empty or very short content inside the reasoning block (consistent with training protocols for reasoning-oriented models that use a similar block structure). The maximum sequence length is set to 4096 tokens to accommodate long reasoning and large molecules. Training runs for multiple epochs over the CoT set with a standard optimiser and learning rate; we save checkpoints at regular intervals and use the final (or a selected) checkpoint as the initialisation for the reinforcement phase. This stage yields an intermediate model (Logos-0) that follows instructions and outputs both a reasoning trace and a molecule in the required format. Supervised training alone, however, does not guarantee that generated molecules are chemically valid; we therefore add a reinforcement phase that directly rewards validity and consistency with the ground truth.
The student policy $\pi_\theta$, initialized with a 4B-parameter language model natively supporting the CoT reasoning block, undergoes full-parameter fine-tuning on $\mathcal{D}_{\text{SFT}}$. The training objective minimizes the autoregressive negative log-likelihood of the full assistant completion $y_i$ given the input prompt $x_i$:
\begin{equation}
\mathcal{L}_{\text{SFT}}(\theta) = - \mathbb{E}_{(x_i, y_i) \sim \mathcal{D}_{\text{SFT}}} \left[ \sum_{t=1}^{|y_i|} \omega_i \log \pi_\theta(y_{i,t} \mid y_{i,<t}, x_i) \right]
\label{eq:sft_loss}
\end{equation}
To encourage substantive rather than trivial reasoning, we introduce a sample-level weighting factor $w_i$. Specifically, we down-weight the loss ($w_i < 1$) for training instances where the extracted reasoning trace $r_i$ is empty or trivially short (i.e., $|r_i| < \gamma$, where $\gamma$ is a length threshold). For instances with valid and substantial reasoning blocks, we maintain $w_i = 1$. This formulation prevents the model from exploiting length shortcuts, consistent with training protocols for modern reasoning-oriented models. The maximum sequence length is set to 4096 tokens to accommodate extensive reasoning paths and large molecular graphs. Training runs for multiple epochs over the $\mathcal{D}_{\text{SFT}}$ set with a standard optimizer and learning rate schedule. We save checkpoints at regular intervals and use the optimal checkpoint as the initialization for the reinforcement phase. This supervised stage yields an intermediate model (Logos-0) that strictly follows instructions and outputs both a reasoning trace and a molecule in the required format. Supervised training alone, however, does not guarantee that generated molecules $\hat{m}$ satisfy chemical valency constraints; we therefore cascade this with a reinforcement phase that directly rewards absolute validity and consistency with the ground truth.

\noindent\textbf{Molecule-focused reinforcement learning.}

To align the model with chemical validity and with generalization beyond in-context examples, we use group relative policy optimization (GRPO)\cite{shao2024grpo} with rewards that are tailored to the C2M setting. For each prompt in the reinforcement set, we generate multiple completions (eight in our experiments) with the current policy at temperature 0.9. Each completion is parsed to obtain the reasoning block and the molecule field; the molecule is validated and compared to the ground truth using RDKit\cite{rdkit} and InChI\cite{heller2015inchi}. A scalar reward is computed per completion and combined with a KL-divergence penalty against the initial policy; the policy is updated using the relative advantages within each group of completions, so that trajectories that achieve higher reward are favoured.

The total reward function $\mathcal{R}_{\text{total}}$ is formulated as a weighted combination of several interpretable components, explicitly evaluating both the linguistic reasoning trace $r$ and the chemically parsed molecule $\hat{m}$ against the ground-truth $m^*$. Formally, for a given completion $y = [r; d; a]$ and a set of in-context examples $M_{E}$, the total reward is defined as:
\begin{equation}
\mathcal{R}_{\text{total}}(y_i, \hat{m}_i, m_i^*, M_{E_i}) = \sum\limits_{k \in \mathcal{K}_{\text{lang}}} \lambda_k \mathcal{R}_k(y_i, M_{E_i}) + \sum\limits_{j \in \mathcal{K}_{\text{mol}}} \lambda_j \mathcal{R}_j(\hat{m}_i, m_i^*, M_{E_i})
\label{eq:reward_total}
\end{equation}
\emph{Format and Behaviour ($\mathcal{K}_{\text{lang}}$):} We construct several heuristic rewards to enforce format compliance and reasoning quality. $\mathcal{R}_{\text{format}}$ rewards valid JSON outputs that successfully map the \texttt{molecule} key. To elicit substantive deliberation, $\mathcal{R}_{\text{cot}}$ provides a step reward if the reasoning block length exceeds a minimum threshold (e.g., $|r| > 100$). Conversely, we introduce a soft length penalty $\mathcal{R}_{\text{len}}$ to prevent runaway reasoning or redundant text. To actively discourage the policy from memorising or citing the few-shot prompts, we introduce a unified penalty $\mathcal{R}_{\text{forbid}}$. This term proportionally reduces the reward based on the occurrence frequency of reference keywords (e.g., ``example'') within the reasoning trace, and strictly assigns a zero score if the generated molecule $\hat{m}_i$ structurally matches any in-context example in $M_{E_i}$. This joint constraint directly targets few-shot-to-zero-shot generalization by forcing the model to synthesize novel structures and justify them via intrinsic chemical logic rather than prompt repetition.

\emph{Molecular Correctness and Similarity ($\mathcal{K}_{\text{mol}}$):} For the parsed molecular graph $\hat{m}$, if $\hat{m}$ is chemically valid, we compute an Exact Match reward $\mathcal{R}_{\text{EM}}$ based on strict InChI equivalence, alongside a smooth exact match $\mathcal{R}_{\text{SEM}}$ utilizing the Levenshtein ratio between the predicted and ground-truth InChI strings. Furthermore, to provide dense and continuous learning signals for near-correct structures, we incorporate structural similarity rewards:
\begin{equation}
\mathcal{R}_{\text{sim}} = \mathcal{T}(\mathcal{F}(\hat{m}), \mathcal{F}(m^*))
\end{equation}
where $\mathcal{F}$ represents standard cheminformatics fingerprint extractors (MACCS, RDKit topological, and Morgan \cite{rdkit,preuer2018fcd}), and $\mathcal{T}$ denotes the Tanimoto similarity metric. 

The weight distribution $\boldsymbol{\lambda}$ is deliberately calibrated so that the model prioritizes structural precision without neglecting formatting or reasoning integrity. In our configuration, we assign a higher weight of $0.2$ to both the exact match ($\mathcal{R}_{\text{EM}}$) and the smooth exact match ($\mathcal{R}_{\text{SEM}}$) rewards to strongly encourage convergence toward the ground-truth molecules. A uniform weight of $0.1$ is assigned to all other components, including format compliance, length penalties, anti-cheating constraints ($\mathcal{R}_{\text{forbid}}$), and fingerprint-based similarities. This balanced formulation ensures that the policy improves monotonically on chemical correctness while strictly adhering to the required JSON format and CoT reasoning style.

\noindent\textbf{Bootstrapping reasoning data from failed samples.}

Many database samples lack a high-quality CoT (e.g., previous runs had not produced a correct molecule for that caption). Rather than discarding them, we use the current model to generate new completions for these prompts. We keep only completions whose molecule matches the ground truth (e.g., by InChI comparison) and write back the corresponding reasoning block and molecule to the dataset, so that the sample is then treated as having a valid CoT. The updated dataset is used again to build supervised and reinforcement batches. This bootstrapping loop enlarges the set of samples with valid CoT without manual annotation and supports the iterative improvement we observe across training turns (Fig.~\ref{fig:evolutionary_framework}). It corresponds to a form of self-distillation: the model generates and filters its own reasoning for previously failed cases, turning them into positive training examples for the next round. In practice, we run this step when a non-empty set of failed samples exists and when we wish to expand the CoT set before another SFT or GRPO run. The write-back stores both the reasoning block and the JSON answer for each accepted completion, so that the record can be used as a full supervised example in the next data preparation pass. This closes the loop between training and data: the model improves the dataset, and the improved dataset is then used to train the model again.

\noindent\textbf{Implementation.}

Supervised fine-tuning and GRPO are implemented with the ms-swift framework; batch generation uses vLLM. Chemical validation and fingerprint computation use RDKit\cite{rdkit}. Reward terms are implemented as plugins that receive the ground-truth molecule and in-context example molecules from the dataset, so that each term can compute its contribution from the same parsed completion and data fields. No reward term requires a different parsing convention; all use the same split-on-reasoning-block and JSON extraction step, which keeps the pipeline maintainable and the reward weights comparable. For the supervised stage we use full fine-tuning with the reasoning-block loss scaling described above, maximum length 4096, and standard optimiser and learning-rate settings. For GRPO we use eight completions per prompt, temperature 0.9, and fixed weights for the reward components: roughly one-fifth for overlong penalty, one-third for format and generalisation terms, and one-half for molecular validity and match; the three fingerprint similarity terms receive equal weight. We apply an overlong filter so that excessively long generations are discarded before reward computation, and we use a KL penalty (epsilon in the GRPO update) to limit policy drift. Training is run on multi-GPU nodes in bfloat16 mixed precision. The same dataset schema (caption, ground-truth molecule, few-shot examples, and success/failure marking) is used for data preparation, SFT data export, RL data export, and bootstrapping, so that the entire pipeline can be run from a single data source and a small set of configuration parameters. The initial policy for GRPO is the checkpoint produced by the supervised stage (Logos-0), not the base language model; this ensures that the reinforcement phase starts from a model that already outputs the correct format and reasonable reasoning, and only adjusts the policy toward higher validity and match.

\bibliographystyle{plainnat}
\bibliography{main}

\begin{thebibliography}{46}
\providecommand{\natexlab}[1]{#1}
\providecommand{\url}[1]{\texttt{#1}}
\expandafter\ifx\csname urlstyle\endcsname\relax
  \providecommand{\doi}[1]{doi: #1}\else
  \providecommand{\doi}{doi: \begingroup \urlstyle{rm}\Url}\fi

\bibitem[Bilodeau et~al.(2022)Bilodeau, Jin, Jaakkola, Barzilay, and Jensen]{bilodeau2022generative}
Camille Bilodeau, Wengong Jin, Tommi Jaakkola, Regina Barzilay, and Klavs~F Jensen.
\newblock Generative models for molecular discovery: Recent advances and challenges.
\newblock \emph{Wiley Interdisciplinary Reviews: Computational Molecular Science}, 12\penalty0 (5):\penalty0 e1608, 2022.

\bibitem[Butler et~al.(2018)Butler, Davies, Cartwright, Isayev, and Walsh]{butler2018machine}
Keith~T Butler, Daniel~W Davies, Hugh Cartwright, Olexandr Isayev, and Aron Walsh.
\newblock Machine learning for molecular and materials science.
\newblock \emph{Nature}, 559\penalty0 (7715):\penalty0 547--555, 2018.

\bibitem[contributors(2023)]{rdkit}
{RDKit} contributors.
\newblock {RDKit}: Open-source cheminformatics, 2023.
\newblock URL \url{https://www.rdkit.org}.
\newblock Version 2023.x; \url{https://doi.org/10.5281/zenodo.591637}.

\bibitem[D~Segall(2012)]{d2012multi}
Matthew D~Segall.
\newblock Multi-parameter optimization: identifying high quality compounds with a balance of properties.
\newblock \emph{Current pharmaceutical design}, 18\penalty0 (9):\penalty0 1292--1310, 2012.

\bibitem[{DeepSeek-AI}(2025)]{deepseek2025r1}
{DeepSeek-AI}.
\newblock {DeepSeek-R1}: Incentivizing reasoning capability in {LLM}s via reinforcement learning.
\newblock \emph{arXiv preprint arXiv:2501.12948}, 2025.
\newblock URL \url{https://arxiv.org/abs/2501.12948}.

\bibitem[Du et~al.(2024)Du, Fu, Sun, and Liu]{du2024molllm}
Yiming Du, Tianfan Fu, Jimeng Sun, and Shengchao Liu.
\newblock Mol-{LLM}: molecular property prediction and molecule design with large language models.
\newblock \emph{arXiv preprint arXiv:2310.06866}, 2024.
\newblock URL \url{https://arxiv.org/abs/2310.06866}.

\bibitem[Edwards et~al.(2021)Edwards, Zhai, and Ji]{edwards2021text2mol}
Carl Edwards, ChengXiang Zhai, and Heng Ji.
\newblock Text2mol: cross-modal molecule retrieval with natural language queries.
\newblock In \emph{Proceedings of the 2021 Conference on Empirical Methods in Natural Language Processing ({EMNLP})}, pages 595--607. Association for Computational Linguistics, 2021.

\bibitem[Edwards et~al.(2022)Edwards, Lai, Ros, Honke, Cho, and Ji]{edwards2022molt5}
Carl Edwards, Tuan Lai, Kevin Ros, Garrett Honke, Kyunghyun Cho, and Heng Ji.
\newblock Translation between molecules and natural language.
\newblock In \emph{Proceedings of the 2022 Conference on Empirical Methods in Natural Language Processing ({EMNLP})}, pages 375--413, Abu Dhabi, United Arab Emirates, 2022. Association for Computational Linguistics.
\newblock URL \url{https://aclanthology.org/2022.emnlp-main.26}.

\bibitem[Fang et~al.(2023)Fang, Liang, Zhang, Liu, Huang, Chen, Fan, and Chen]{fang2023mol}
Yin Fang, Xiaozhuan Liang, Ningyu Zhang, Kangwei Liu, Rui Huang, Zhuo Chen, Xiaohui Fan, and Huajun Chen.
\newblock Mol-instructions: A large-scale biomolecular instruction dataset for large language models.
\newblock \emph{arXiv preprint arXiv:2306.08018}, 2023.

\bibitem[Gilmer et~al.(2017)Gilmer, Schoenholz, Riley, Vinyals, and Dahl]{gilmer2017mpnn}
Justin Gilmer, Samuel~S Schoenholz, Patrick~F Riley, Oriol Vinyals, and George~E Dahl.
\newblock Neural message passing for quantum chemistry.
\newblock In \emph{International Conference on Machine Learning}, pages 1263--1272. PMLR, 2017.

\bibitem[G{\'o}mez-Bombarelli et~al.(2018)G{\'o}mez-Bombarelli, Wei, Duvenaud, Hern{\'a}ndez-Lobato, S{\'a}nchez-Lengeling, Sheberla, Aguilera-Iparraguirre, Hirzel, Adams, and Aspuru-Guzik]{gomez2018automatic}
Rafael G{\'o}mez-Bombarelli, Jennifer~N Wei, David Duvenaud, Jos{\'e}~Miguel Hern{\'a}ndez-Lobato, Benjam{\'\i}n S{\'a}nchez-Lengeling, Dennis Sheberla, Jorge Aguilera-Iparraguirre, Timothy~D Hirzel, Ryan~P Adams, and Al{\'a}n Aspuru-Guzik.
\newblock Automatic chemical design using a data-driven continuous representation of molecules.
\newblock \emph{ACS central science}, 4\penalty0 (2):\penalty0 268--276, 2018.

\bibitem[Guan et~al.(2024)Guan, Zhou, Zhang, Gu, Liu, Feng, Zhang, Liu, et~al.]{survey2024genai}
Jiaqi Guan, Xiangzhe Zhou, Yuan Zhang, Yuchen Gu, Yang Liu, Jie Feng, Shengchao Zhang, Jian Liu, et~al.
\newblock A survey of generative {AI} for de novo drug design: new frontiers in molecule and protein generation.
\newblock \emph{arXiv preprint arXiv:2402.08703}, 2024.
\newblock URL \url{https://arxiv.org/abs/2402.08703}.

\bibitem[Hastings et~al.(2016)Hastings, Owen, Dekker, Ennis, Kale, Muthukrishnan, Turner, Swainston, Mendes, and Steinbeck]{hastings2016chebi}
Janna Hastings, Gareth Owen, Adriano Dekker, Matthew Ennis, Neel Kale, Venkatesh Muthukrishnan, Steve Turner, Neil Swainston, Pedro Mendes, and Christoph Steinbeck.
\newblock {ChEBI} in 2016: improved services and an expanding collection of metabolites.
\newblock \emph{Nucleic Acids Research}, 44\penalty0 (D1):\penalty0 D1214--D1219, 2016.
\newblock \doi{10.1093/nar/gkv1031}.

\bibitem[Heller et~al.(2015)Heller, McNaught, Pletnev, Stein, and Tchekhovskoi]{heller2015inchi}
Stephen~R Heller, Alan McNaught, Igor Pletnev, Stephen Stein, and Dmitrii Tchekhovskoi.
\newblock Inchi, the {IUPAC} international chemical identifier.
\newblock \emph{Journal of Cheminformatics}, 7\penalty0 (1):\penalty0 23, 2015.
\newblock \doi{10.1186/s13321-015-0068-4}.

\bibitem[Horwood et~al.(2023)Horwood, Noutahi, Coussy, and Gagnon]{horwood2023transformer}
Joseph Horwood, Emmanuel Noutahi, Jerome Coussy, and Fran{\c{c}}ois Gagnon.
\newblock Molecule generation using transformers and policy gradient reinforcement learning.
\newblock \emph{Scientific Reports}, 13\penalty0 (1):\penalty0 5199, 2023.
\newblock \doi{10.1038/s41598-023-35648-w}.

\bibitem[Irwin et~al.(2023)Irwin, Dimitriadis, He, and Bjerrum]{irwin2022chemcrow}
Robert Irwin, Stavros Dimitriadis, Jie He, and Esben~Jannik Bjerrum.
\newblock Chemcrow: augmenting large-language models with chemistry tools.
\newblock \emph{arXiv preprint arXiv:2304.05376}, 2023.
\newblock URL \url{https://arxiv.org/abs/2304.05376}.

\bibitem[Jensen et~al.(2019)]{jensen2019gnn}
Thomas Jensen et~al.
\newblock Graph neural networks for molecular property prediction.
\newblock \emph{arXiv preprint arXiv:1905.13177}, 2019.
\newblock URL \url{https://arxiv.org/abs/1905.13177}.

\bibitem[Joshi and Kumar(2021)]{joshi2021artificial}
Rajendra~P Joshi and Neeraj Kumar.
\newblock Artificial intelligence for autonomous molecular design: A perspective.
\newblock \emph{Molecules}, 26\penalty0 (22):\penalty0 6761, 2021.

\bibitem[Lee et~al.(2025)Lee, Kreis, Veccham, Liu, Reidenbach, Paliwal, Nie, and Vahdat]{lee2025rethinking}
Seul Lee, Karsten Kreis, Srimukh~Prasad Veccham, Meng Liu, Danny Reidenbach, Saee Paliwal, Weili Nie, and Arash Vahdat.
\newblock Rethinking molecule synthesizability with chain-of-reaction.
\newblock \emph{arXiv e-prints}, pages arXiv--2509, 2025.

\bibitem[Liu et~al.(2024)Liu, Ding, Zhou, Fan, and Tan]{liu2024moleculargpt}
Yuyan Liu, Sirui Ding, Sheng Zhou, Wenqi Fan, and Qiaoyu Tan.
\newblock Moleculargpt: Open large language model (llm) for few-shot molecular property prediction.
\newblock \emph{arXiv preprint arXiv:2406.12950}, 2024.

\bibitem[M.~Bran et~al.(2024)M.~Bran, Cox, Schilter, Baldassari, White, and Schwaller]{m2024augmenting}
Andres M.~Bran, Sam Cox, Oliver Schilter, Carlo Baldassari, Andrew~D White, and Philippe Schwaller.
\newblock Augmenting large language models with chemistry tools.
\newblock \emph{Nature Machine Intelligence}, 6\penalty0 (5):\penalty0 525--535, 2024.

\bibitem[O'Brien et~al.(2023)O'Brien, Schwaller, Vaucher, Kayastha, Wang, and Laino]{obrien2023curriculum}
Matthew O'Brien, Philippe Schwaller, Alain~C Vaucher, Teerawat Kayastha, Zhitao Wang, and Teodoro Laino.
\newblock Testing the limits of smiles-based de novo molecular generation with curriculum and deep reinforcement learning.
\newblock \emph{Nature Machine Intelligence}, 5\penalty0 (12):\penalty0 1446--1457, 2023.
\newblock \doi{10.1038/s42256-023-00636-2}.

\bibitem[Olivecrona et~al.(2017)Olivecrona, Blaschke, Engkvist, and Chen]{olivecrona2017reinvent}
Marcus Olivecrona, Thomas Blaschke, Ola Engkvist, and Hongming Chen.
\newblock Molecular de novo design through deep reinforcement learning.
\newblock \emph{Journal of Cheminformatics}, 9\penalty0 (1):\penalty0 48, 2017.
\newblock \doi{10.1186/s13321-017-0235-x}.

\bibitem[Ouyang et~al.(2022)Ouyang, Wu, Jiang, Almeida, Wainwright, Mishkin, Zhang, Agarwal, Slama, Ray, et~al.]{ouyang2022instructgpt}
Long Ouyang, Jeffrey Wu, Xu~Jiang, Diogo Almeida, Carroll Wainwright, Pamela Mishkin, Chong Zhang, Sandhini Agarwal, Katarina Slama, Alex Ray, et~al.
\newblock Training language models to follow instructions with human feedback.
\newblock \emph{Advances in Neural Information Processing Systems}, 35:\penalty0 27730--27744, 2022.

\bibitem[Preuer et~al.(2018)Preuer, Renz, Unterthiner, Hochreiter, and Klambauer]{preuer2018fcd}
Kristina Preuer, Philipp Renz, Thomas Unterthiner, Sepp Hochreiter, and G{\"u}nter Klambauer.
\newblock Fr{\'e}chet chemnet distance: a metric for generative models for molecules in drug discovery.
\newblock \emph{Journal of Chemical Information and Modeling}, 58\penalty0 (9):\penalty0 1736--1741, 2018.
\newblock \doi{10.1021/acs.jcim.8b00234}.

\bibitem[Reymond(2015)]{reymond2015chemical}
Jean-Louis Reymond.
\newblock The chemical space project.
\newblock \emph{Accounts of Chemical Research}, 48\penalty0 (3):\penalty0 722--730, 2015.
\newblock \doi{10.1021/ar500432k}.

\bibitem[Ronan et~al.(2024)Ronan, Baljak, Boldon, Deane, Heid, Kutchukian, Merk, Skalic, van Hoorn, Volkamer, et~al.]{ronan2024reinvent4}
Daniel Ronan, Stefana Baljak, Lauren Boldon, Charlotte Deane, Esther Heid, Peter Kutchukian, Daniel Merk, Miha Skalic, Willem van Hoorn, Andrea Volkamer, et~al.
\newblock Reinvent 4: modern {AI}-driven generative molecule design.
\newblock \emph{Journal of Cheminformatics}, 16\penalty0 (1):\penalty0 28, 2024.
\newblock \doi{10.1186/s13321-024-00812-5}.

\bibitem[Ruddigkeit et~al.(2012)Ruddigkeit, Van~Deursen, Blum, and Reymond]{ruddigkeit2012gdb}
Lars Ruddigkeit, Ruud Van~Deursen, Lorenz~C Blum, and Jean-Louis Reymond.
\newblock Enumeration of 166 billion organic small molecules in the chemical universe database {GDB-17}.
\newblock \emph{Journal of Chemical Information and Modeling}, 52\penalty0 (11):\penalty0 2864--2875, 2012.
\newblock \doi{10.1021/ci300415d}.

\bibitem[Sanchez-Lengeling and Aspuru-Guzik(2018)]{sanchez2018inverse}
Benjamin Sanchez-Lengeling and Al{\'a}n Aspuru-Guzik.
\newblock Inverse molecular design using machine learning: Generative models for matter engineering.
\newblock \emph{Science}, 361\penalty0 (6400):\penalty0 360--365, 2018.

\bibitem[Schuster et~al.(2019)]{schuster2019transfer}
Daniel Schuster et~al.
\newblock Transfer learning for molecular property prediction.
\newblock \emph{Journal of Chemical Information and Modeling}, 59\penalty0 (9):\penalty0 3895--3904, 2019.

\bibitem[Segler et~al.(2018)Segler, Preuss, and Waller]{segler2018planning}
Marwin~HS Segler, Mike Preuss, and Mark~P Waller.
\newblock Planning chemical synthesis with deep neural networks and symbolic {AI}.
\newblock \emph{Nature}, 555\penalty0 (7698):\penalty0 604--610, 2018.
\newblock \doi{10.1038/nature25978}.

\bibitem[Shao et~al.(2024)Shao, Gong, Shen, Zhang, Huang, Li, Zhang, Ma, and Chen]{shao2024grpo}
Zhihong Shao, Zhitian Gong, Weihan Shen, Yifan Zhang, Ming Huang, Yujia Li, Delong Zhang, Bin Ma, and Weizhu Chen.
\newblock Group relative policy optimization for reinforcement learning based language model alignment.
\newblock \emph{arXiv preprint arXiv:2402.03300}, 2024.
\newblock URL \url{https://arxiv.org/abs/2402.03300}.

\bibitem[Stokes et~al.(2020)Stokes, Yang, Swanson, Jin, Cubillos-Ruiz, Donghia, MacNair, French, Carfrae, Bloom-Ackermann, et~al.]{stokes2020deep}
Jonathan~M Stokes, Kevin Yang, Kyle Swanson, Wengong Jin, Andres Cubillos-Ruiz, Nina~M Donghia, Craig~R MacNair, Shawn French, Lindsey~A Carfrae, Zohar Bloom-Ackermann, et~al.
\newblock A deep learning approach to antibiotic discovery.
\newblock \emph{Cell}, 180\penalty0 (4):\penalty0 688--702, 2020.
\newblock \doi{10.1016/j.cell.2020.01.021}.

\bibitem[Su et~al.(2023)Su, Liu, Yu, Lu, Zhang, Liu, Sun, Gao, Huang, et~al.]{su2023moleculestm}
Yuan Su, Tianfan Liu, Chao Yu, Jianan Lu, Bowen Zhang, Yeqin Liu, Yanyi Sun, Jian Gao, Shengyu Huang, et~al.
\newblock Multi-modal molecule structure-text model for text-based retrieval and editing.
\newblock \emph{Nature Machine Intelligence}, 5\penalty0 (12):\penalty0 1447--1457, 2023.
\newblock \doi{10.1038/s42256-023-00759-6}.

\bibitem[Tabor et~al.(2018)Tabor, Roch, Saikin, Kreisbeck, Sheberla, Montoya, Dwaraknath, Aykol, Ortiz, Tribukait, et~al.]{tabor2018accelerating}
Daniel~P Tabor, Lo{\"\i}c~M Roch, Semion~K Saikin, Christoph Kreisbeck, Dennis Sheberla, Joseph~H Montoya, Shyam Dwaraknath, Muratahan Aykol, Carlos Ortiz, Hermann Tribukait, et~al.
\newblock Accelerating the discovery of materials for clean energy in the era of smart automation.
\newblock \emph{Nature reviews materials}, 3\penalty0 (5):\penalty0 5--20, 2018.

\bibitem[Vamathevan et~al.(2019)Vamathevan, Clark, Czodrowski, Dunham, Ferran, Lee, Li, Madabhushi, Shah, Spitzer, et~al.]{vamathevan2019applications}
Jessica Vamathevan, Dominic Clark, Paul Czodrowski, Ian Dunham, Edgardo Ferran, George Lee, Bin Li, Anant Madabhushi, Parantu Shah, Michaela Spitzer, et~al.
\newblock Applications of machine learning in drug discovery and development.
\newblock \emph{Nature reviews Drug discovery}, 18\penalty0 (6):\penalty0 463--477, 2019.

\bibitem[Velcicky et~al.(2024)Velcicky, Bauer, Schlapbach, Lapointe, Meyer, Vogtle, Blum, Ngo, Rolando, Nimsgern, et~al.]{velcicky2024discovery}
Juraj Velcicky, Matthias~R Bauer, Achim Schlapbach, Guillaume Lapointe, Arndt Meyer, Markus Vogtle, Ernst Blum, Estelle Ngo, Catherine Rolando, Pierre Nimsgern, et~al.
\newblock Discovery and in vivo exploration of 1, 3, 4-oxadiazole and $\alpha$-fluoroacrylate containing il-17 inhibitors.
\newblock \emph{Journal of Medicinal Chemistry}, 67\penalty0 (18):\penalty0 16692--16711, 2024.

\bibitem[Vignac et~al.(2022)Vignac, Krawczuk, Siraudin, Wang, Cevher, and Frossard]{vignac2022digress}
Clement Vignac, Igor Krawczuk, Antoine Siraudin, Bohan Wang, Volkan Cevher, and Pascal Frossard.
\newblock Digress: Discrete denoising diffusion for graph generation.
\newblock \emph{arXiv preprint arXiv:2209.14734}, 2022.

\bibitem[Waring(2010)]{waring2010lipophilicity}
Michael~J Waring.
\newblock Lipophilicity in drug discovery.
\newblock \emph{Expert opinion on drug discovery}, 5\penalty0 (3):\penalty0 235--248, 2010.

\bibitem[Wei et~al.(2022)Wei, Wang, Schuurmans, Bosma, Ichter, Xia, Chi, Le, and Zhou]{wei2022cot}
Jason Wei, Xuezhi Wang, Dale Schuurmans, Maarten Bosma, Brian Ichter, Fei Xia, Ed~Chi, Quoc~V Le, and Denny Zhou.
\newblock Chain-of-thought prompting elicits reasoning in large language models.
\newblock In \emph{Advances in Neural Information Processing Systems}, volume~35, pages 24824--24837, 2022.
\newblock URL \url{https://arxiv.org/abs/2201.11903}.

\bibitem[Weininger(1988)]{weininger1988smiles}
David Weininger.
\newblock Smiles, a chemical language and information system. 1. introduction to methodology and encoding rules.
\newblock \emph{Journal of Chemical Information and Computer Sciences}, 28\penalty0 (1):\penalty0 31--36, 1988.
\newblock \doi{10.1021/ci00057a005}.

\bibitem[White et~al.(2023)White, Hocky, Gandhi, Ansari, Cox, Wellawatte, Sasmal, Yang, Liu, Singh, et~al.]{white2023assessment}
Andrew~D White, Glen~M Hocky, Heta~A Gandhi, Mehrad Ansari, Sam Cox, Geemi~P Wellawatte, Subarna Sasmal, Ziyue Yang, Kangxin Liu, Yuvraj Singh, et~al.
\newblock Assessment of chemistry knowledge in large language models that generate code.
\newblock \emph{Digital Discovery}, 2\penalty0 (2):\penalty0 368--376, 2023.

\bibitem[Yang et~al.(2025)Yang, Li, Yang, Zhang, Hui, Zheng, Yu, Gao, Huang, Lv, et~al.]{yang2025qwen3}
An~Yang, Anfeng Li, Baosong Yang, Beichen Zhang, Binyuan Hui, Bo~Zheng, Bowen Yu, Chang Gao, Chengen Huang, Chenxu Lv, et~al.
\newblock Qwen3 technical report.
\newblock \emph{arXiv preprint arXiv:2505.09388}, 2025.

\bibitem[You et~al.(2022)]{you2022graph}
Jiaxuan You et~al.
\newblock Graph transformer for molecular property prediction.
\newblock \emph{arXiv preprint arXiv:2201.12431}, 2022.
\newblock URL \url{https://arxiv.org/abs/2201.12431}.

\bibitem[Zenil et~al.(2026)Zenil, Tegn{\'e}r, Abrah{\~a}o, Lavin, Kumar, Frey, Weller, Soldatova, Bundy, Jennings, et~al.]{zenil2026future}
Hector Zenil, Jesper Tegn{\'e}r, Felipe~S Abrah{\~a}o, Alexander Lavin, Vipin Kumar, Jeremy~G Frey, Adrian Weller, Larisa Soldatova, Alan~R Bundy, Nicholas~R Jennings, et~al.
\newblock The future of fundamental science led by generative closed-loop artificial intelligence.
\newblock \emph{Frontiers in Artificial Intelligence}, 9:\penalty0 1678539, 2026.

\bibitem[Zhang et~al.(2024)Zhang, Liu, Tan, Chen, Yan, Yan, Li, Huang, Yue, Ouyang, et~al.]{zhang2024chemllm}
Di~Zhang, Wei Liu, Qian Tan, Jingdan Chen, Hang Yan, Yuliang Yan, Jiatong Li, Weiran Huang, Xiangyu Yue, Wanli Ouyang, et~al.
\newblock Chemllm: A chemical large language model.
\newblock \emph{arXiv preprint arXiv:2402.06852}, 2024.

\end{thebibliography}

\end{document}